\documentclass[conference]{IEEEtran}

\usepackage{amsmath}                
\usepackage{amssymb}
\usepackage{amsfonts}
\usepackage{balance}
\usepackage{graphicx}               
 
\usepackage[english]{babel}  
\usepackage[utf8]{inputenc}     
\usepackage[T1]{fontenc}
\usepackage{ae,aecompl}         
 
\usepackage{microtype}              
 
\usepackage{hyperref}
\usepackage{subcaption}
\usepackage{tabularx}
\usepackage{booktabs}
\usepackage{enumerate}              
\usepackage{algorithm2e}            
\usepackage{todonotes}              
\usepackage{dirtree}                    

\usepackage{hhline}
\usepackage{caption}
\usepackage{lmodern}
\usepackage{nccmath}
\usepackage{scalerel,stackengine}
\stackMath
\newcommand\reallywidehat[1]{%
    \savestack{\tmpbox}{\stretchto{%
            \scaleto{%
                \scalerel*[\widthof{\ensuremath{#1}}]{\kern.1pt\mathchar"0362\kern.1pt}%
                {\rule{0ex}{\textheight}}
            }{\textheight}%
        }{2.4ex}}%
    \stackon[-6.9pt]{#1}{\tmpbox}%
}
\usepackage{soul}
\usepackage{color}

\usepackage{transparent}
 
\definecolor{reviewer1}{rgb}{0.235, 0.471, 0.847}
\definecolor{reviewer2}{rgb}{0.416, 0.659, 0.31}
\definecolor{reviewer3}{rgb}{0.902, 0.568, 0.22}
\definecolor{redc}{rgb}{1.0, 0, 0}

\usepackage{calrsfs}
\DeclareMathAlphabet{\pazocal}{OMS}{zplm}{m}{n}

\usepackage{graphicx,stackengine,scalerel}
 
\graphicspath{{./figures/}}
\usepackage{float}
\usepackage{multicol}

\begin{document}
\title{Autonomous Multi-Robot Exploration Strategies for 3D Environments with Fire Detection Capabilities}
\author{\IEEEauthorblockN{Ankit Shaw}
\IEEEauthorblockA{ \textit{College of Engineering} \\
    \textit{University of Washington, Seattle, USA} \\
\textit{ankit25@uw.edu}}
}
\maketitle
\begin{abstract}
\noindent
    { This paper presents a comprehensive overview of exploration strategies utilized in both 2D and 3D environments, focusing on autonomous multi-robot systems designed for building exploration and fire detection. We explore the limitations of traditional algorithms that rely on prior knowledge and predefined maps, emphasizing the challenges faced when environments undergo changes that invalidate these maps. Our modular approach integrates localization, mapping, and trajectory planning to facilitate effective exploration using an OctoMap framework generated from point cloud data. The exploration strategy incorporates obstacle avoidance through potential fields, ensuring safe navigation in dynamic settings. Additionally, I propose future research directions, including decentralized map creation, coordinated exploration among unmanned aerial vehicles (UAVs), and adaptations to time-varying environments. This work serves as a foundation for advancing coordinated multi-robot exploration algorithms, enhancing their applicability in real-world scenarios.}
\end{abstract}
\section{Introduction}
\label{sec:intro}

The application of multi-robot systems to solving core robotics problems has garnered significant attention in recent decades. A typical example is the deployment of robot teams for exploring unknown environments, which is particularly relevant in scenarios such as search and rescue missions \cite{Murphy2004}, automated cleaning tasks \cite{Endres}, warehouse management \cite{Wurman2008}, and planetary exploration \cite{Mataric2001}. As multi-robot systems become increasingly integrated into various sectors, the development of efficient coordination algorithms has become essential.

For autonomous navigation in unknown areas, robots require maps of the environment, a process known as \textit{autonomous exploration} \cite{Julia2012}. A team of robots, compared to a single robot, can accomplish tasks more quickly by distributing the workload, fusing sensor data to create a more accurate representation of the environment, and compensating for individual sensor uncertainties \cite{Dias2000}, \cite{Wurm2008}. Effective multi-robot coordination offers several benefits, such as reduced exploration time, increased robustness, and higher-quality maps. Moreover, coordinated multi-robot systems can handle tasks that may be unmanageable for a single robot due to physical constraints or limited sensing capabilities \cite{Dias2006}.

This paper aims to review state-of-the-art 2D and 3D exploration strategies for multi-robot systems, providing insights into various methodologies used for autonomous navigation and mapping. We present a detailed discussion on decentralized 2D exploration strategies, followed by an introduction to a 3D autonomous building exploration framework designed specifically for fire detection and monitoring.

\section{2D Exploration Strategies}
\label{sec:2d_exploration}

Successful exploration in robotics relies heavily on accurate localization and mapping techniques. These maps are commonly represented as occupancy grid maps \cite{Moravec}, where the environment is discretized into a grid of cells. Each cell is categorized as \textit{free}, \textit{occupied}, or \textit{unknown} based on sensor measurements. One of the pioneering exploration strategies, \textit{frontier-based exploration}, was introduced by Yamauchi \cite{Yamauchi1997}, where frontiers are defined as the boundary cells between known and unknown regions. These frontiers serve as targets for the robots, and the nearest frontier is selected for exploration \cite{Julia2012}.

Exploration strategies can be broadly classified into \textit{centralized} and \textit{decentralized} approaches. In centralized strategies, a leader or central planner is responsible for task assignment to each robot in the team \cite{Dias2000}. This approach can generate globally optimal exploration plans but suffers from communication overhead, limited scalability, and potential single-point failures \cite{Julia2012}. On the other hand, decentralized approaches allow each robot to independently choose its actions based on its local knowledge and perception of the environment \cite{Yan2013}. While these methods may not always provide globally optimal solutions, they offer improved flexibility, robustness, and scalability \cite{Zlot2002}.

\subsection{Nearest Frontier Approach}
\label{subsec:nearest_frontier}

The \textit{nearest frontier approach} is one of the most straightforward strategies for autonomous exploration. Robots identify the nearest frontier cell and plan the shortest path to reach it \cite{Yamauchi1997}. While this approach is computationally efficient and easy to implement, it lacks explicit coordination mechanisms among multiple robots, which can lead to suboptimal performance, such as multiple robots targeting the same frontier. Several extensions have been proposed to address these issues, including the use of a bidding process for task allocation \cite{Burgard2005} and dense frontier detection methods to speed up the exploration process \cite{Orsulic2019}.

\subsection{Cost-Utility Approach}
\label{subsec:cost_utility}

The \textit{cost-utility} approach is another prominent strategy that attempts to balance the expected information gain from exploring a target region with the cost of reaching that region \cite{GonzlezBaos2002}. The benefit of a candidate cell is typically evaluated as a linear combination of the utility and the cost, as defined below:

\begin{equation}
B_{CU}(a) = U(a) - \lambda_{CU}C(a),
\end{equation}

where \( U(a) \) represents the expected utility of exploring cell \( a \), and \( C(a) \) denotes the cost associated with reaching the cell. The parameter \( \lambda_{CU} \) is a scaling factor that balances the trade-off between exploration gain and traversal cost. An example of this approach is Burgard's method, which employs a utility function that considers both the information gain and the travel cost. The frontier assignment to robots is optimized using the Hungarian method to achieve a balanced and efficient allocation \cite{Burgard2005}.

\section{3D Exploration Strategies} \label{sec:3d_exploration}

\begin{figure}[h]
	\centering
	\includegraphics[width=1.0\columnwidth]{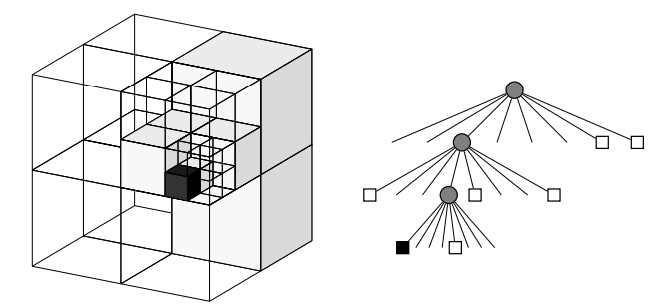}	
	\caption{Illustration of the octree data-structure. Left: Example of an octree storing free voxels (shaded white) and occupied voxels (black). Right: The corresponding tree representation (Source: \cite{Wurm2012}).}
	\label{fig:octomap}
\end{figure}

\begin{figure}[h]
	\centering
	\includegraphics[width=1.0\columnwidth]{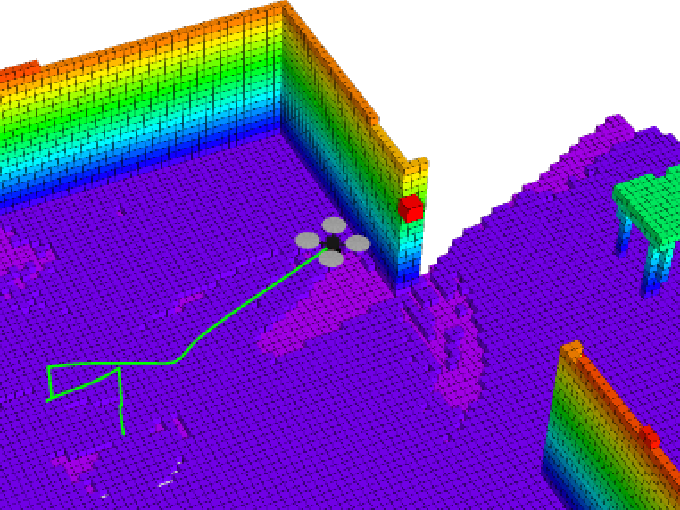}	
	\caption{An unmanned aerial vehicle (UAV) exploring a 3D environment using OctoMap. The colored voxels represent the 3D OctoMap, and the green lines show the UAV’s exploration trajectory (Source: \cite{Wang2019}).}
	\label{fig:octomapanddrone}
\end{figure}

Robots operating autonomously in a three-dimensional (3D) environment require a robust model to represent regions that are safe to traverse and obstacles to avoid. This section outlines several exploration strategies adopted for navigating 3D spaces. These strategies often rely on data structures such as OctoMap to effectively represent both occupied and free spaces. The OctoMap framework \cite{Wurm2012} discretizes the environment by dividing it into cubic volumes, known as \textit{voxels}, and maintains the probability of each voxel being occupied. It stores only essential data in a tree-based representation, significantly reducing memory usage. The OctoMap structure, along with a tree-based representation of free and occupied voxels, is illustrated in Fig. \ref{fig:octomap}. An example of a drone-based exploration strategy using OctoMap is shown in Fig. \ref{fig:octomapanddrone}.

In contrast to 2D exploration and mapping strategies, 3D exploration requires considerably more computational power and memory resources. Strategies for 3D exploration can be implemented using a single robot or multiple robots working in coordination. The following subsections discuss various strategies for single-robot and multi-robot 3D exploration.

\subsection{Single Robot Exploration} \label{subsec:single_robot_exploration}
Various autonomous 3D exploration strategies for a single robot are described in the literature. Joho et al. \cite{Joho2007} extended well-known 2D exploration strategies to 3D environments, utilizing multi-level surface maps and cost functions that account for expected information gain and travel cost. Bachrach et al. \cite{Bachrach2009} developed a 2D frontier-based exploration algorithm, with a fixed altitude for a quadrotor helicopter, enabling the helicopter to autonomously explore and map unstructured indoor environments.

In another work, Dornhege and Kleiner \cite{Dornhege2013} adapted the frontier-based exploration method to 3D spaces. However, their approach demands high computational power and is limited to small workspaces. On the other hand, Maurovic et al. \cite{Maurovic2014} proposed a 3D exploration strategy for a mobile robot equipped with a 3D laser scanner. Their method includes an online room-detection algorithm that explores environments room-by-room, keeping memory and computational requirements low.

A next-best-view approach for building a 3D model of a real object without any a priori knowledge was described by Vasquez-Gomez et al. \cite{VasquezGomez2014}. The algorithm optimizes each view to reconstruct arbitrary objects while dealing with sensor uncertainty. Bircher et al. \cite{Bircher2016} presented a novel path planning algorithm for 3D exploration. The proposed planner computes an online tree to select the best branch, considering the amount of unmapped space that can be explored. This method is capable of running in real-time on a robot with limited resources.

Baiming et al. \cite{Baiming2018} introduced a trajectory planning algorithm that uses target points for exploring unknown space. Their method plans long-term target points and local trajectories in real-time, enabling the robot to explore unknown environments while avoiding obstacles. Senarathne et al. \cite{Senarathne2016} proposed an alternative approach based on surface frontier voxels. This strategy seeks to expand mapped surfaces rather than merely reducing unmapped voxels.

Wang et al. \cite{Wang2018} tackled the problem of autonomous exploration in unknown indoor environments using mutual information. The authors proposed a sampling method that extracts random sensing patches in free space, extended to informative locations to collect information. They combined this with Gaussian Markov Random Fields (GMRF) to model the distribution of mutual information, proposing a utility function that balances path cost and information gain.

\subsection{Multi-Robot Exploration} \label{subsec:multi_robot_exploration}
In multi-robot systems, collaboration and information sharing play crucial roles. Several strategies have been proposed to enable effective cooperation among robots. Priyasad et al. \cite{Priyasad2018} developed a point cloud-based algorithm that enables exploration in environments with highly inaccurate prior knowledge. Their method utilizes depth images to construct a local map and searches for uncharted areas using a breadth-first approach. Zhu et al. \cite{Zhu2015} employed a frontier-based strategy for autonomous exploration using a micro aerial vehicle (MAV) equipped with 3D sensors. They utilized OctoMap to build real-time maps during exploration.

Vutetakis \cite{Vutetakis2019} proposed a strategy for inspecting critical infrastructure using MAVs. The strategy focuses on the autonomous exploration and coverage of unknown structures to develop high-fidelity 3D models. To minimize accumulative data errors, the algorithm plans loop closures during exploration.

Rocha et al. \cite{Rocha2005} addressed 3D mapping using multiple robots with cubic cells for information storage. A frontier-based exploration strategy is employed, where initial maps are updated with new information obtained through sensing. Robots share their local maps to avoid redundancy and optimize exploration efficiency. The process continues until the entire area is explored and mapped.

\begin{figure}[h]
	\centering
	\includegraphics[width=1.0\columnwidth]{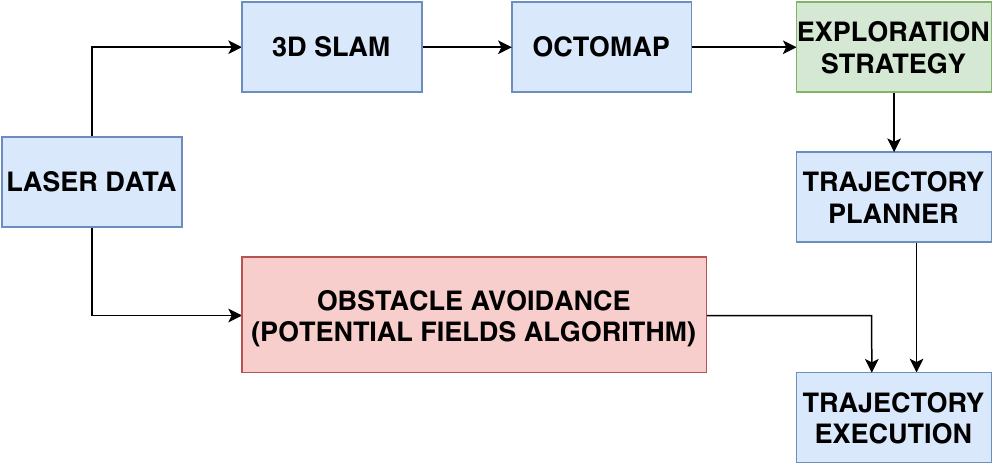}	
	\caption{Diagram of the implemented 3D exploration strategy.}
	\label{fig:3D_strategy}
\end{figure}

Wang et al. \cite{Wang2019} tackled the problem of 3D exploration using aerial vehicles with limited flight endurance. Their approach uses an information potential field-based method that considers both travel cost and information gain. The algorithm selects the next-best-view point based on a multi-objective function, balancing the information gathered and the travel path cost. An example of the 3D exploration strategy is shown in Fig. \ref{fig:3D_strategy}.

The schematic diagram of the implemented exploration and mapping process is presented in Fig. \ref{fig:3D_strategy}. The exploration process involves localization, mapping, trajectory planning, and OctoMap generation. OctoMap is created using Google Cartographer's \textit{submap point cloud}. Cartographer includes global and local SLAM subsystems: local SLAM generates submaps of the region, while global SLAM stitches the submaps together. The resulting OctoMap is then used for trajectory planning and execution, as shown in Fig. \ref{fig:rviz_gazebo}.

\begin{figure}[h]
	\centering
	\includegraphics[width=1.0\columnwidth]{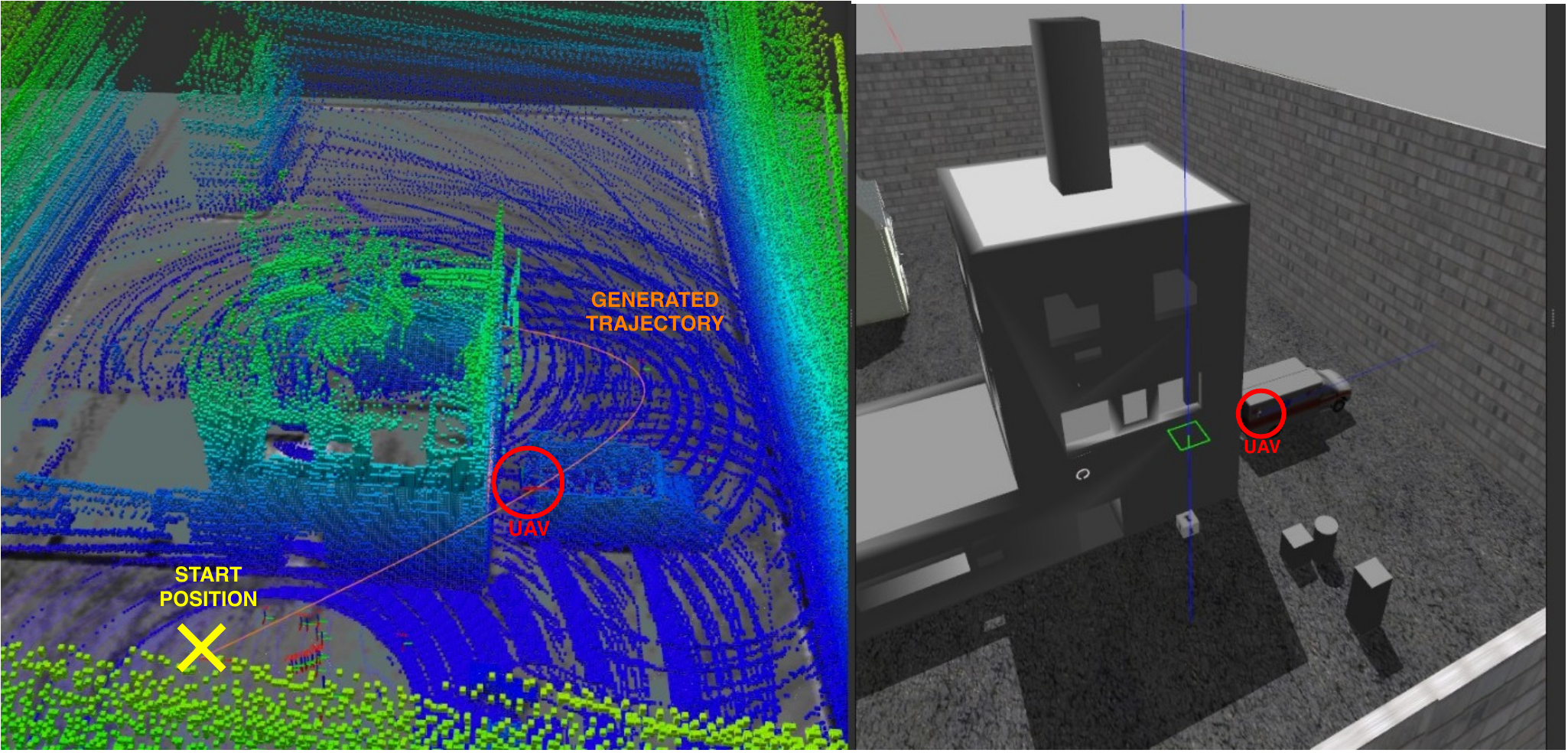}	
	\caption{Execution of the generated trajectory using the OctoMap representation.}
	\label{fig:rviz_gazebo}
\end{figure}

\begin{figure}[h]
	\centering
	\includegraphics[width=1.0\columnwidth]{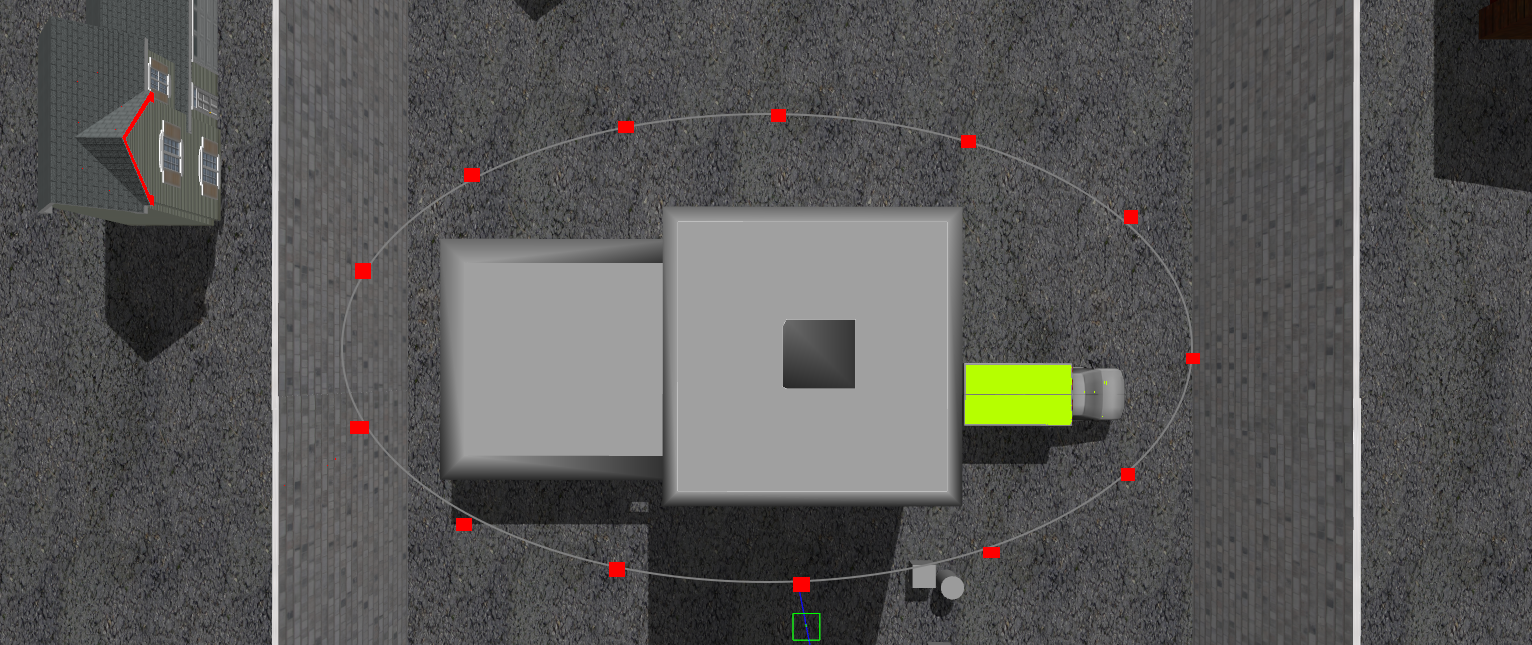}	
	\caption{Ellipse-shaped target points generated from the GPS location of the building center.}
	\label{fig:building}
\end{figure}

The target points for trajectory generation are currently obtained from the Global Positioning System (GPS) location of the building. Fig. \ref{fig:building} shows ellipse-shaped target points generated based on the GPS coordinates of the building center.

An obstacle avoidance algorithm based on potential fields is also considered in our work. Since the map is created online and trajectory planning does not involve replanning during execution, the generated trajectory points must be locally validated and corrected if necessary. The robot's laser data is sent to the obstacle avoidance algorithm. If a robot's current position is too close to a static obstacle or if a dynamic obstacle is detected, the obstacle avoidance algorithm is activated, and the generated trajectory is modified accordingly.

\section{Conclusion and Future Work} \label{sec:conclusion}

In this paper, I have presented a comprehensive overview of existing exploration strategies for multi-robot systems in both 2D and 3D spaces. The discussed approaches demonstrate the potential benefits and challenges of using multi-robot systems for autonomous exploration and mapping. Our work also introduces a modular and decentralized strategy for effective exploration and mapping in multi-robot scenarios, emphasizing the importance of autonomy and collaboration among the robots.

In terms of future research directions, several avenues can be explored to enhance the capabilities of multi-robot exploration systems. One promising area is the development of algorithms that enable decentralized map creation and management during exploration. This would allow each robot to maintain and update its own map while ensuring global consistency through occasional inter-robot communication. Such a decentralized approach can significantly reduce the dependency on a central coordinator, improving scalability and robustness of the system.

Given the limitations of existing strategies, another potential research direction is addressing the challenges associated with restricted communication ranges. Multi-robot systems often operate in environments where maintaining communication links can be difficult due to obstacles, signal interference, or long distances. Future work could focus on optimizing exploration algorithms to cope with these communication constraints by implementing techniques such as multi-hop communication, message prioritization, or dynamically adaptive communication protocols.

Additionally, improving the efficiency of frontier detection and filtering mechanisms remains an open problem. Combining frontier-based exploration with information-theoretic approaches or machine learning methods can help robots identify high-quality frontiers more effectively, thereby reducing redundant exploration and improving overall mapping performance.

Regarding 3D exploration, this paper provides an overview of current strategies and presents a solution for autonomous exploration of 3D environments using a single robot without prior knowledge of the map. Expanding this approach to a multi-robot or multi-UAV (Unmanned Aerial Vehicles) system could further enhance the efficiency and effectiveness of 3D mapping. Coordinated exploration strategies, such as task allocation, formation control, or shared path planning, can be developed to leverage the unique capabilities of each robot, allowing them to work collaboratively to cover large and complex environments.

Map sharing is another area that warrants further investigation. Enabling robots to share partial or complete maps can drastically improve the collective understanding of the environment and reduce exploration time. Research could focus on developing robust map-merging techniques that can handle discrepancies between locally generated maps and overcome uncertainties in robot positioning.

Lastly, it is essential to consider scenarios where robots may encounter failures, such as sensor malfunctions, mechanical breakdowns, or navigation errors. Developing fault-tolerant exploration strategies that allow the remaining robots to continue their tasks despite such failures will increase the overall resilience of the system. Furthermore, exploring strategies for dynamic and time-varying environments can enable robots to adapt to changes in the environment, such as moving obstacles, evolving terrain, or varying environmental conditions.

In conclusion, this work serves as a stepping stone for new research in the field of coordinated multi-robot exploration algorithms and their decentralization. We believe that the insights provided in this paper will inspire future research efforts aimed at improving the autonomy, efficiency, and robustness of multi-robot exploration systems in both 2D and 3D environments.

\begin{figure}[h]
    \centering
    \includegraphics[width=1.0\columnwidth]{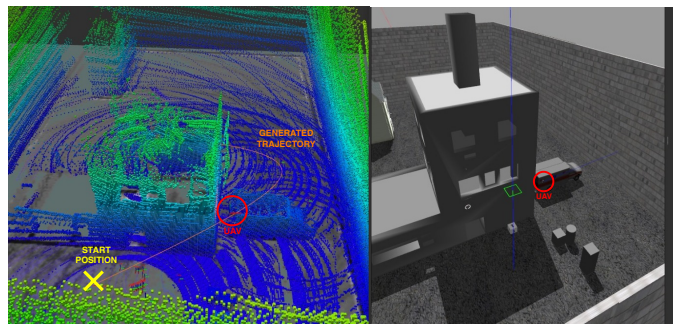}
    \caption{Execution of the generated trajectory in a simulated environment using Gazebo and visualization in RViz. This figure illustrates the trajectory followed by the robot in a 3D exploration task.}
    \label{fig:rviz_gazebo}
\end{figure}

\bibliographystyle{IEEEtran}



\end{document}